\documentclass{article}
\usepackage{graphicx} 
\usepackage[final]{ARLET_2025}
\usepackage[utf8]{inputenc} 
\usepackage[T1]{fontenc}    
\usepackage{hyperref}       
\usepackage{url}            
\usepackage{booktabs}       
\usepackage{amsfonts}       
\usepackage{nicefrac}       
\usepackage{microtype}      
\usepackage{xcolor}         
\usepackage[dvipsnames]{xcolor}
\usepackage{xspace}
\usepackage{amsmath}
\usepackage{amsthm}
\usepackage{bbm}
\usepackage{natbib}
\usepackage{algpseudocode}

\makeatletter
\newcommand\appendix@section[1]{%
  \refstepcounter{section}%
  \orig@section*{Appendix \@Alph\c@section: #1}%
  \addcontentsline{toc}{section}{Appendix \@Alph\c@section: #1}%
}
\let\orig@section\section
\g@addto@macro\appendix{\let\section\appendix@section}
\makeatother

\usepackage[ruled, vlined, linesnumbered]{algorithm2e}
\SetKwInOut{Input}{Input}
\SetKwInOut{Require}{Require}
\SetKwInOut{Initialize}{Initialize}

\newcommand{\rqucb}{\textsc{RestartQ-UCB}\xspace}
\newcommand{\randomized}{\textsc{RANDOMIZEDQ}\xspace}
\newcommand{\partialr}{PartialRestarts\xspace}
\newcommand{\adaptiver}{AdaptiveRestarts\xspace}
\newcommand{\selectiver}{SelectiveRestarts\xspace}

\newcommand{\selectiverandomized}{\textsc{SelectiveRANDOMIZEDQ}\xspace}
\newcommand{\adaparrqucb}{\textsc{AdaParRestartQ}-UCB\xspace}

\newcommand{\set}[1]{\{ #1 \}}
\newcommand{\defeq}{\stackrel{\text{def}}{=}}
\DeclareMathOperator*{\argmax}{arg\,max}

\DeclareMathOperator*{\softmax}{softmax}
\DeclareMathOperator{\sign}{sign}

\newtheorem{manuallemmainner}{Lemma}
\newenvironment{manuallemma}[1]{%
  \renewcommand\themanuallemmainner{#1}%
  \manuallemmainner
}{\endmanuallemmainner}

\newcommand*\samethanks[1][\value{footnote}]{\footnotemark[#1]}

\title{Efficient Restarts in Non-Stationary Model-Free Reinforcement Learning}
\author{
  Hiroshi Nonaka \thanks{These authors contributed equally to this work} \\
  Soka University of America \\
  \texttt{hnonaka@soka.edu} \\
  \And
  Simon Ambrozak \samethanks \\
  University of Maryland, College Park \\
  \texttt{sambroza@umd.edu} \\
  \And
  Sofia R. Miskala-Dinc \thanks{These authors contributed equally to this work} \\
  University of Maryland, College Park \\
  \texttt{smiskala@umd.edu} \\
  \And
  Amedeo Ercole \samethanks \\
  University of Maryland, College Park \\
  \texttt{aercole@umd.edu} \\
  \And
  Aviva Prins \\
  University of Maryland, College Park \\
  \texttt{aviva@umd.edu} \\
}\date{September 2025}

\begin{document}

\maketitle

\begin{abstract}
    In this work, we propose three efficient restart paradigms for model-free non-stationary reinforcement learning (RL). We identify two core issues with the restart design of \citet{mao2022modelfreenonstationaryrlnearoptimal}'s \rqucb algorithm: (1) complete forgetting, where all the information learned about an environment is lost after a restart, and (2) scheduled restarts, in which restarts occur only at predefined timings, regardless of the incompatibility of the policy with the current environment dynamics. We introduce three approaches, which we call \textit{partial}, \textit{adaptive}, and \textit{selective} restarts to modify the algorithms \rqucb and \randomized \citep{wang2025provablyefficientagilerandomized}. We find near-optimal empirical performance in multiple different environments, decreasing dynamic regret by up to $91$\% relative to \rqucb.
\end{abstract}

\section{Introduction}
Reinforcement learning (RL) is a computational approach to solving interactive problems, such as crop management, inventory control, and board games \citep{gautron2022reinforcementlearningforcropmanagement, lu2023onlineoptimizationbaseddecisionsupport, 
mahajan2025comparativeanalysismultiagentreinforcement, 
tao2023optimizingcropmanagementreinforcement, mao2022modelfreenonstationaryrlnearoptimal, silver2016masteringthegameofgo, silver2017masteringchessshogiselfplay}. While conventional RL assumes stationarity, many real-life settings present more complex dynamics called non-stationarity, in which reward functions and transitions change over time. To model these real-world scenarios, we focus on reinforcement learning in episodic non-stationary Markov decision process (MDP) structures. These environments pose unique challenges for learning.

RL for non-stationary environments is difficult because algorithms must manage two key trade-offs: \textit{exploration versus exploitation} and \textit{remembering versus forgetting}. The exploration-exploitation trade-off is inherent to RL.
This trade-off becomes even more challenging in non-stationary environments. Since the environment is changing, an RL algorithm must explore more in order to account for change, but it loses out on reward by not exploiting during that time. In regards to remembering versus forgetting, an agent must decide what data to keep and what to discard, since previously acquired information about the environment may no longer be valid.

\citet{mao2022modelfreenonstationaryrlnearoptimal} introduce a model-free algorithm called \rqucb, which manages these trade-offs in such a way that promises near-optimal performance. It uses an optimism term to encourage exploration and performs an occasional \textit{restart} that forgets all previously seen observations. Despite an impressive asymptotic performance guarantee, we find that fully erasing all learned data at scheduled intervals does not best utilize the information that the agent has observed.

To address these inefficiencies, we discuss three paradigms, which we call \textit{partial}, \textit{adaptive}, and \textit{selective} restarts. Partial restarts address the issue of complete forgetting by resetting the $Q$-table to a tighter upper bound than the loose bound that is commonly used. Adaptive restarts identify when restarts are most needed, rather than relying on a fixed hyperparameter. Finally, selective restarts combine these two approaches and reset only a subset of $Q$-table entries to further accelerate convergence to a new optimum.

We evaluate our approaches by comparing cumulative reward in two environments: a new randomized environment that we call RandomMDP and Bidirectional Diabolical Combination Locks (BDCL) \citep{agarwal2020pc-pg}. Our empirical results are impressive. We observe a $74$\% and $91$\% decrease in dynamic regret in RandomMDP and BDCL environments, respectively. The latter result is particularly impressive because the BDCL environment is designed to be difficult to explore. These results demonstrate the potential of our frameworks to overcome the restart inefficiencies. We hope that by adding our restart modifications to a theoretically robust algorithm, we can achieve near-optimal performance in practice while maintaining the spirit of \rqucb's asymptotic guarantees.

\section{Related work}
Conventional RL studies consider stationary environments. Those foundational works include value-based methods, such as \textsc{Q-learning}  \citep{watkins1992qlearning} and \textsc{deep Q-learning} \citep{mnih2013playingatarideepreinforcement}, and policy-gradient and actor-critic methods, which include \textsc{REINFORCE} \citep{williams1992simple} and \textsc{Actor-Critic} \citep{konda1999actor-criticalgorithms}. These algorithms demonstrate good performance in various stationary MDPs. Nonetheless, many real-world scenarios involve non-stationary dynamics, which poses a fundamental limitation to the RL algorithms \citep{dasilva2006dealingwith, gautron2022reinforcementlearningforcropmanagement, zhou2024nonstationaryreinforcementlearninglinear, mao2022modelfreenonstationaryrlnearoptimal}. The central challenge, therefore, is to enable RL agents to continuously adapt to these unpredictable and time-varying conditions. Many works introduce model-based RL approaches to solve the non-stationarity issues. \citet{gajane2019variationalregretboundsreinforcement} propose Variation-aware UCRL, a variant of the UCRL algorithm that restarts according to a schedule dependent on the total variation in the MDP. \citet{cheung2020reinforcementlearningnonstationarymarkov} introduce the Sliding Window UCRL2 with Confidence Widening (\textsc{SWUCRL2-CW}) algorithm, which uniquely addresses challenges posed by conventional optimistic exploration techniques in non-stationary MDPs by incorporating additional optimism through a confidence widening technique. \citet{dasilva2006dealingwith} approach this problem by introducing a multiple-model approach called \textsc{RL-CD}, where it evaluates the prediction quality of several partial models and incrementally builds new ones as needed, selecting the most appropriate one when a context change is detected.

While most approaches have been model-based, they typically suffer from time and space complexity \citep{jin2018isQlearningprovablyefficient, mao2022modelfreenonstationaryrlnearoptimal, zhang2020optimalmodelfreereinforcementlearning}. This has motivated a growing interest in model-free approaches, which offer advantages in terms of online applicability and flexibility. \citet{mao2022modelfreenonstationaryrlnearoptimal} developed \rqucb, which is a pioneering model-free non-stationary RL for episodic MDPs, achieving competitive dynamic regret by adopting a simple restarting strategy and incorporating an extra optimism term.

\section{Preliminaries}

We consider a non-stationary finite-horizon episodic Markovian setting. An instance of a Markov decision process (MDP) is given by $\mathcal{M} = (\mathcal{S}, \mathcal{A}, H, P, r)$, where $\mathcal{S}$ is a finite set of states, 
$\mathcal{A}$ is a finite set of actions,
$P = \set{P^m_h}_{h\in [H], m \in [M]}$ is a set of transition kernels, and $r = \set{r^m_h}_{h\in[H], m\in[M]}$ is a set of reward functions. The setting contains $M$ episodes, each of length $H$. We define $T$ as the total number of steps in the entire horizon, where $T=MH$ and $M, H, T \in \mathbb{N}$. For $N \in \mathbb{N}$, we use the notation convention $[N] \defeq [1, 2, \dots, N]$. We denote $S = |\mathcal{S}|$ and $A = |\mathcal{A}|$. An agent observes a state $s^m_h \in \mathcal{S}$ and takes an action $a^m_h \in \mathcal{A}$ by following a policy function. The environment returns a reward $r^m_h(s^m_h, a^m_h) \in [0, 1]$. The environment then transitions to a new state $s^m_{h+1} \sim P^m_h(\cdot \mid s^m_h, a^m_h)$. Since the environment is non-stationary, $P^m_h$ and $r^m_h$ may vary over time with respect to $m$ and $h$.

The degree of non-stationarity of a given MDP is quantified via \textit{variation budget}. This is defined as the sum of the supremum distances between rewards or transition probabilities across two consecutive episodes:
\begin{align}\label{eq:Delta_r}
    \Delta_r & \defeq \sum^{M-1}_{m=1} \sum^H_{h=1} \sup_{s, a} \left| r^m_h(s, a) - r^{m+1}_h(s, a) \right|
\end{align}
\begin{align}\label{eq:Delta_p}
    \Delta_p & \defeq \sum^{M-1}_{m=1} \sum^H_{h=1} \sup_{s, a} || P^m_h(\cdot \mid s, a) - P^{m+1}_h(\cdot \mid s, a) ||_1,
\end{align}
where $|| \cdot ||_1$ denotes $L^1$-norm. $\Delta_r$ and $\Delta_p$ represent the total amount of change in reward functions and transitions episode to episode over the entire horizon, thus indicating the extent of non-stationarity in the environment.

The goal of the RL agent is to solve for a policy of actions $\pi$, defined by the collection of functions $\pi^m_h: \mathcal{S} \rightarrow \mathcal{A}$. Thus, $a^m_h = \pi^m_h(s^m_h)$. Under this finite setting, there exists an optimal policy $\pi^\star$ that maximizes total reward. A state value function $V^{m, \pi}_h: \mathcal{S} \rightarrow \mathbb{R}$ returns a scalar quantifying the value of a state at episode $m$ and step $h$ under a policy $\pi$:
\[
V^{m, \pi}_h \defeq \mathbb{E} \left[ \sum^H_{h'=h} r^m_{h'}(s_{h'}, \pi^m_{h'}(s_{h'})) \mid s_{h} = s , s_{h'+1} \sim P^m_h(\cdot \mid s_{h'}, a_{h'}) \right]
\]

Likewise, an action value function $Q^m_h: \mathcal{S} \times \mathcal{A} \rightarrow \mathbb{R}$ returns the value of a state-action pair under a policy:
\[
Q^m_h(s, a) \defeq r^m_h(s, a) + \mathbb{E} \left[ \sum^H_{h'=h+1} r^m_{h'}(s_{h'}, \pi^m_{h'}(s_{h'})) \mid s_{h} = s, a_h=a, s_{h'+1} \sim P^m_h(\cdot \mid s_{h'}, a_{h'}) \right]
\]
Analogously, the state and action value functions of $\pi^\star$ are given by $V^{m, \star}_h(s) = \sup_{\pi}V^m_h(s)$ and $Q^{m, \star}_h(s, a) = \sup_{\pi}Q^m_h(s, a)$. Note that in this episodic setting we may let $V_{H+1}^{m, \pi}(s) = 0, \forall s \in \mathcal{S}, m \in [M]$.

The goal of the agent is to maximize total cumulative reward. \textit{Dynamic regret} is the cumulative difference between the optimal and policy-based state values per episode:
\[
\mathcal{R}(\pi, M) \defeq \sum^M_{m=1} \left[ V^{m, \star}_1(s^m_1) - V^{m, \pi}_1(s_1^m) \right].
\]
Since dynamic regret measures the amount of reward that the agent missed out on each episode, relative to a (fixed, but possibly unknown) optimal policy, maximizing total cumulative reward is analogous to minimizing dynamic regret.

\subsection{Limitations of existing approaches}\label{sec:existing_limitations}

\rqucb is a $Q$-learning algorithm that utilizes upper confidence bounds and scheduled restarts that reset the learned $Q$-values every time the end of an \textit{epoch} is reached. An epoch is a sequential group of episodes, and any learning in one epoch does not affect learning in another because \rqucb is completely restarted when an epoch ends. The number of epochs is set to $D = S^{-\frac{1}{3}}A^{-\frac{1}{3}}\Delta^{\frac{2}{3}}H^{-\frac{2}{3}}T^{\frac{1}{3}}$, and the number of episodes in each epoch is $K=\left\lceil \frac{M}{D} \right\rceil$. The algorithm is reproduced in Algorithm \ref{alg:rqucb}. Although \rqucb has a near-optimal upper bound on dynamic regret $\tilde{O}\left(S^\frac{1}{3} A^\frac{1}{3} (\Delta_r + \Delta_p)^\frac{1}{3} HT^\frac{2}{3}\right)$, our empirical analysis of its performance reveals a significant gap between theory and practice. Most MDPs will have some exploitable structure that we seek to take advantage of while not losing performance guarantees in the worst-case scenarios.

\RestyleAlgo{ruled}
\begin{algorithm}[H] \label{alg:rqucb}
\caption{RestartQ-UCB (Hoeffding), \citet{mao2022modelfreenonstationaryrlnearoptimal}}
\For{epoch $d \gets 1$ to $D$}
    {\textbf{Initialize:} $V_h(s) \gets H - h + 1$, $Q_h(s,a) \gets H - h + 1$,
    $N_h(s,a) \gets 0$, $\check{N}_h(s,a) \gets 0$, 
    $\check{r}_h(s,a) \gets 0$, $\check{v}_h(s,a) \gets 0$, 
    $\check{\mu}_h(s,a) \gets 0$, $\check{\sigma}_h(s,a) \gets 0$,
    $\mu^{\text{ref}}_h(s,a) \gets 0$, $\sigma^{\text{ref}}_h(s,a) \gets 0$, 
    $V^{\text{ref}}_h(s) \gets H$, for all $(s,a,h) \in \mathcal{S} \times \mathcal{A} \times [H]$.
    
    \For{episode $k \gets (d-1)K + 1$ to $\min\{dK, M\}$}
        {\textbf{observe} $s_1$;
        
        \For{step $h \gets 1$ to $H$}
        {
            \textbf{Take} action $a_h \gets \arg\max_a Q_h(s_h,a)$;
            \textbf{Receive} reward $R_h(s_h,a_h)$ and observe $s_{h+1}$
            $
                \check{r}_h(s_h,a_h) \gets \check{r}_h(s_h,a_h) + R_h(s_h,a_h),
            $
            $
                \check{v}_h(s_h,a_h) \gets \check{v}_h(s_h,a_h) + V_{h+1}(s_{h+1}).
            $
            $
            N_h(s_h,a_h) \gets N_h(s_h,a_h) + 1, \quad
            \check{N}_h(s_h,a_h) \gets \check{N}_h(s_h,a_h) + 1
            $
            
            \If{$N_h(s_h,a_h) \in \mathcal{L}$}{
                $
                    b_h \gets \sqrt{\tfrac{H^2}{\check{N}_h(s_h,a_h)} \, \iota}
                         + \sqrt{\tfrac{1}{\check{N}_h(s_h,a_h)} \, \iota}, \quad
                    b_\Delta \gets \Delta_r^{(d)} + H \cdot \Delta_p^{(d)}
                $
                
                $
                    Q_h(s_h,a_h) \gets \min \Bigl\{
                       \tfrac{\check{r}_h(s_h,a_h)}{\check{N}_h(s_h,a_h)}
                       + \tfrac{\check{v}_h(s_h,a_h)}{\check{N}_h(s_h,a_h)}
                       + b_h + 2b_\Delta, 
                       Q_h(s_h,a_h) \Bigr\}
                $
                
                $
                    V_h(s_h) \gets \max_a Q_h(s_h,a)
                $
                
                $
                    \check{N}_h(s_h,a_h) \gets 0, \quad
                    \check{r}_h(s_h,a_h) \gets 0, \quad
                    \check{v}_h(s_h,a_h) \gets 0
                $
                }
        }
        }
    }
\end{algorithm}

We identify two sources of learning inefficiency in \rqucb that stem from its restart design. (1) \textbf{Complete forgetting}: \rqucb initializes the $Q$-table to the theoretical maximum value at every restart after an epoch, which requires learning from scratch every time and inflates dynamic regret. (2) \textbf{Scheduled restarts}: \rqucb restarts only at the predetermined timings regardless of whether the learned policy needs a restart, that is, whether the policy is incompatible with the current dynamics of the environment or not. To deal with these practical issues, we introduce three restart algorithmic frameworks in the following section, offering more granular control and potentially superior performance by minimizing unnecessary exploration and computational overhead.

\section{Proposed approach: towards more efficient convergence}

In this section, we introduce partial, adaptive, and selective restarts, which we develop to address the issues of complete forgetting and scheduled restarts.

\subsection{Partial restarts} \label{sec:partial restarts}
When \rqucb performs a restart at the end of an epoch, it fully resets all learned values, and all the information about the environment is forgotten. However, if the agent has knowledge of the environment's variation budgets $\Delta_p$ and $\Delta_r$ (Equations \ref{eq:Delta_r} and \ref{eq:Delta_p}), it is possible to solve for a tighter upper bound. We call this a \textit{partial} restart, because, at a reset, each $Q$-value is raised to some value that is lower than their theoretical maximums.

The goal of a restart in \rqucb is for each $Q$-value in the algorithm's $Q$-table to be larger than its optimal $Q$-value, $Q^\star$. \rqucb achieves this by setting each $Q$-value to the theoretical maximum, $H-h+1$ (line 2 of Algorithm \ref{alg:rqucb}). In this way, the $Q$-values are sure to be greater than $Q^\star$, but generally are much greater than necessary and cause inefficient convergence. Partial restarts use two pieces of information: the learned $Q$-value at the end of the epoch, and the maximum theoretical difference in $Q^\star$-values given $\Delta_r$ and $\Delta_p$. By adding the difference to the learned $Q$-value, we can partially restart it in an efficient way.

$\Delta_p$ and $\Delta_r$ can be used to describe a maximum possible difference in $Q^\star$-values at two different episodes, represented by $Q^{k_2, \star}_h(s,a) - Q^{k_1, \star}_h(s,a)$. Building off of Lemma 1 from \citet{mao2022modelfreenonstationaryrlnearoptimal}, we can show the following:

\begin{manuallemma}{1}\label{lemma1}
    For any triple $(s,a,h)$ and any episodes $k_1, k_2 \in [K]$, it holds that $ \left| Q^{k_2, \star}_h(s,a)  - Q^{k_1, \star}_h(s,a) \right| \leq \Delta_r + \frac{1}{2}\Delta_p\min_{k \in k_1, k_2}\left[ \max_{s,a,h' > h}\left[Q^{k, \star}_h(s,a)\right]\right]$.
\end{manuallemma}

The proof is similar to the method used by \citet{mao2022modelfreenonstationaryrlnearoptimal}, using backwards induction on $h$, the most significant difference is bounding the difference in transition probabilities algebraically instead of using Hölder's inequality. The full proof is provided in Appendix \ref{sec: proof of lemma1}.

\begin{figure}

\centering
\includegraphics[width=0.5\textwidth]{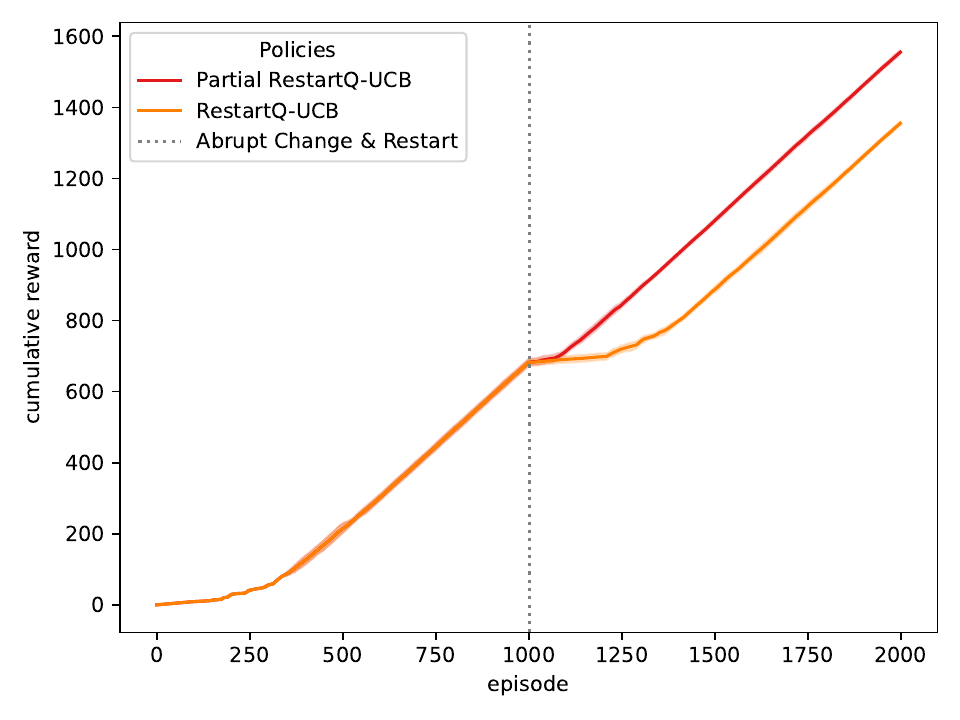}

\caption{This figure compares the impact of a partial restart (red) as opposed to a full restart (orange), when both are positioned to align with an abrupt change in BDCL. After the abrupt change and restarts at episode 1001, partial restarts allow for much faster learning than full restarts.}

\label{fig:BDCL_partial_restart}

\end{figure}

By adding the above bound on the $Q^\star$-value difference to every learned $Q$-value, we can achieve a successful restart while retaining some environmental information given two assumptions: 

\begin{enumerate}

\item Q-UCB learning requires that for all $(s,a,h)$ triples, $Q_h(s,a) \geq Q^\star_h(s,a)$ immediately after the restart, where $Q_h(s,a)$ is the learned $Q$-value for $(s,a,h)$.

\item A learned $Q$-value for a $(s,a,h)$ triple will never go below the lowest optimal $Q$-value for that triple during an epoch. Equivalently: $Q^K_h(s,a) \geq \min_{k\in[K]}Q^{k, \star}_h(s,a)$, where $Q^K_h(s,a)$ is the learned $Q$-value at the end of the epoch, before the restart occurs.
\end{enumerate}
We expect this modification to have the greatest effect in environments with very sparse rewards, such as BDCL, because learned $Q$-values may become very low compared to their theoretical maximums. In such cases, partial restarts allow for much faster learning and convergence to the optimal policy than a full restart. We give an example of this in Figure \ref{fig:BDCL_partial_restart}. In this example, the partial variant recovers faster from an abrupt change in the dynamics of the system. Since the calculations only need to run once per restart, it has a negligible effect on time and space complexity of \rqucb.

\subsection{Adaptive restarts} \label{sec: adaptive restarts}
\rqucb restarts at the beginning of a new epoch, which is calculated using $S$, $A$, $\Delta$, $H$, and $T$. While these timings are chosen to ensure the dynamic regret bound, in practice, scheduled restarts occur at times when a restart is unnecessary. Rather, the agent ought to be using its current policy to maximize reward. To address this problem, we introduce \textit{adaptive} restarts, which detect change in the environment by looking at cumulative reward.

Adaptive restarts estimate how much reward will be gained if a restart happens immediately, and how much reward will be gained if no restart happens. This is done using a sliding window approach over the total reward gained per episode. We note that our use of the term ``sliding window'' is different from \citet{cheung2020reinforcementlearningnonstationarymarkov}. The main idea is that if all of the rewards in the window are summed up, we can keep track of the lowest total reward, highest total reward, and current total reward gained in that window. The lowest total reward will be the reward gained during learning (if our algorithm is running as intended). The highest total reward will be the reward we can gain if we know the environment well, and the current total reward is the reward we are currently gaining.

First, we need to decide the length of the sliding window. Since we want to keep track of how much reward we gain during learning, the window length $W$ should be the number of episodes it took for the agent to begin exploiting the best path it has found. $W$ is computed based on $Q$-table updates while \rqucb begins running.

To find $W$, we keep track of how often the optimal action changes when the agent attempts to update its $Q$-table on line 9 of Algorithm \ref{alg:rqucb}. If an update is triggered (that is, $N_h(s,a) \in \mathcal{L}$) and the optimal action at that state-timestep does not change (i.e., $\argmax_a Q_h(s,a)$ stays the same), we count it as one ``non-update''. Similarly, if the optimal action \textit{does} change, we count it as one ``true-update''. If $H^2$ ``true-updates'' happen before $H^2$ ``non-updates'' occur, then the agent is still learning, and both counters are reset to 0. If $H^2$ "non-updates" happen before $H^2$ ``true-updates'', then learning is considered done, and the window length $W$ is set to the difference between the current episode index and the episode index when the update counters were last reset.

Now, we sum over the first $W$ episodes to get our total reward gained during learning, $r_L$. Every time we reach the end of an episode, we sum over the last $W$ episodes to get our current reward $r_C$, and the highest value of $r_C$ seen is the best reward gained, $r_B$. 

Since we are in a finite-horizon setting, we assume the agent has knowledge of the simulation horizon $T$. Using this, we estimate the total reward we will receive if we \textit{do not} restart: $r_C\frac{T-t}{HW}$, where $t$ is the current timestep, and the reward we will receive if we \textit{do} restart: $r_L + r_B(\frac{T-t}{HW} - 1)$. 

Therefore, if the following equality is true,
\[
r_C\frac{T-t}{HW} < r_L + r_B(\frac{T-t}{HW} - 1)
\]
then a restart occurs. Full pseudocode can be found in Appendix \ref{sec:adaptivealgorithm}.

In Figure \ref{fig:BDCL_ada_restart}, we give an example to demonstrate that adaptive and partial restarts can be used in combination to efficiently overcome the challenges posed a rapidly changing environment. Empirically, \rqucb with adaptive restarts (blue) outperforms scheduled restarts (orange), on the left. This is seen with an even greater effect in the graph on the right, which shows \rqucb with partial and adaptive restarts (purple).

\begin{figure}
\includegraphics[width=0.5\textwidth]{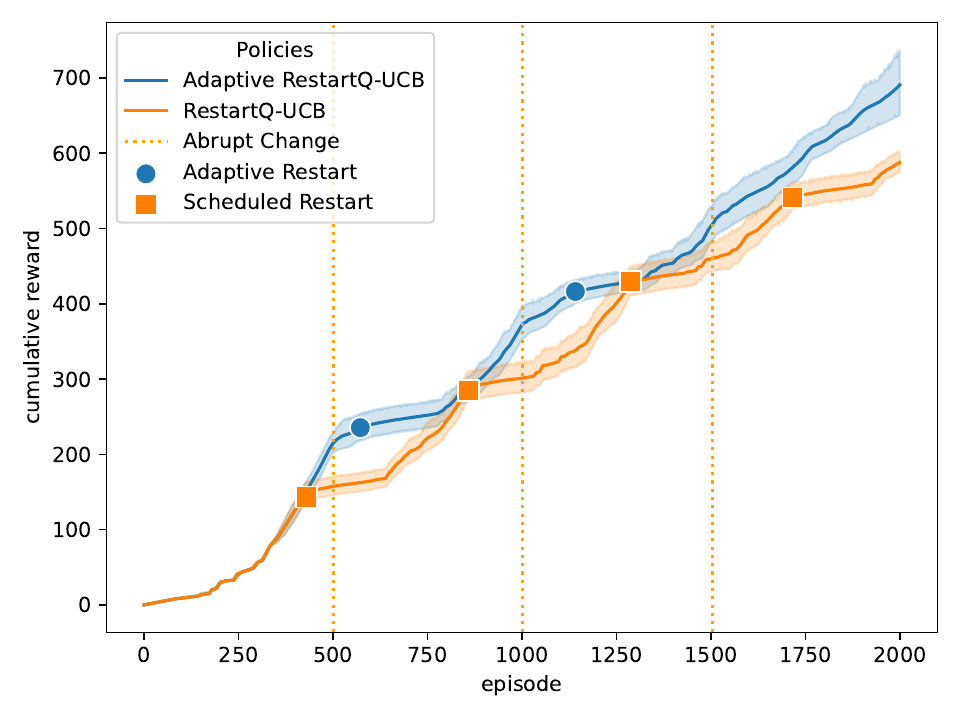}
\includegraphics[width=0.5\textwidth]{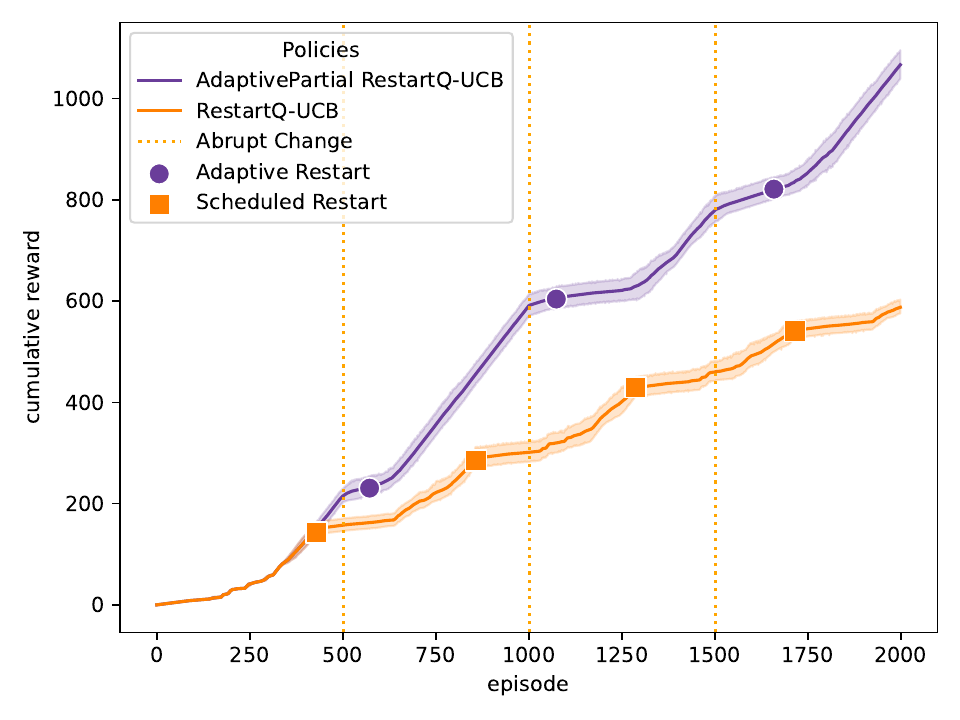}
\caption{This figure demonstrates that adaptive restarts (blue) perform better than scheduled restarts (orange) in BDCL. On the left are adaptive restarts and scheduled restarts, showing that adaptive restarts only occur after each abrupt change and achieve a higher cumulative reward. This effect is further shown on the right, where \rqucb with adaptive and partial restarts (purple) receives nearly twice as much total reward as scheduled, full restarts.}
\label{fig:BDCL_ada_restart}
\end{figure}

Because we are only keeping track of three values: $r_C$, $r_L$, rand $r_B$, and only calculating a running sum of rewards, the time complexity impact is negligible, and the space needed is at worst $O(T)$.

\subsection{Selective restarts}
Finally, we introduce selective restarts, wherein we selectively restart only certain $(s,a)$ positions in the $Q$-table using the upper bound between optimal action values from Lemma \ref{lemma1}. The timing of restarts is also adaptively determined. In this respect, selective restarts are a combined approach of partial and adaptive restarts, but tailored to updating only some entries. 

In our implementation, a selective restart updates $Q_{h'}(s_{h'}, a_{h'})$ associated with the experiential trajectory to $(s, a, h)$:
$\mathcal{T}_h(s, a) \defeq \{ (h', s_{h'}, a_{h'}, s_{h'+1}) \in [ H-h ] \times [S] \times [A] \times [S] \}$ such that $s_{h'+1}=s$ at $h' = h-1$. 

This update rule is based on the upper bound $\beta_h(s, a)$ from Lemma \ref{lemma1}. If $\beta_h(s, a)$ is larger than the absolute difference between the step-wise Bellman update $U^k_h(s, a) = r^k_h(s, a) + V_{h+1}(s_{h+1})$ at the current episode $k$ and $U^{k_0}_h(s, a)$, where $k_0$ is the last episode when $(s, a, h)$ was visited, then the algorithm traces through the trajectory and increment $Q$-values by 
\begin{equation} \label{eqn:deltaQ}
\Delta Q_{h'}(s_{h'}, a_{h'}) = \sign(U^k_h(s, a) - U^{k_0}_h(s, a)) \gamma \frac{1}{H - {h'}}\beta_{h'}(s_{h'}, a_{h'}),
\end{equation}
where $\sign$ returns the sign of an input, and $\gamma$ is a scaling coefficient. We define
$\gamma \defeq \softmax (Q_{h'}(s_{h'}, a_{h'}))$. This $\gamma$ adaptively scales $\beta_{h'}(s_{h'}, a_{h'})$ based on the ratio of $Q$-values along the action dimension, ensuring the update amount associated with actions with a high $Q$-value is weighed more. The scaling coefficient $\frac{1}{H-h'}$ further scales down $\beta_{h'}(s_{h'}, a_{h'})$ based on its step index since state-action pairs in early steps branch out to different trajectories, and thus, get updated more often. We provide pseudocode in Appendix \ref{sec:selectivealgorithm}.

This approach selectively updates specific $Q(s, a, h)$ positions, so it performs the best when the variance of $Q^m_h(s, \cdot)$ is small enough that a series of selective restarts can affect the actions chosen by the policy $\pi^m_h(s) = \argmax_a Q^h_m(s, \cdot)$. Therefore, we may use as a base \textit{any} algorithm that promises quick convergence in a \textit{stationary} environment. Thus, for our empirical results in the following section, we utilize \citet{wang2025provablyefficientagilerandomized}'s \randomized. \randomized is a $Q$-learning-based stationary RL algorithm that adopts the step-wise (\textit{agile}) descents of $Q$-estimates, whereas \rqucb updates $Q$-table at every learning stage $\in \mathcal{L}$, in which the interval of the stages increases exponentially. Instead of UCB algorithms, \randomized uses posterior sampling and $Q$-ensemble methods to address the exploration-exploitation trade-off. We discover that \randomized functions significantly better with selective restarts than \rqucb does, due to the above characteristics. Therefore, in the following experiments, we focus on the performance of \selectiver + \randomized.

The additional time complexity generated by \selectiver at each timestep is $\approx O(B)$, where $B \defeq |\mathcal{T}_h(s, a)|$. Therefore, the total time complexity combined with the base RL algorithm is $\approx O(MH (C_\pi + B))$, where $C_\pi$ is the time complexity of the base RL's policy update function. In practice, $\mathcal{T}_h(s, a)$ is reset at every learning stage, and visitations $(h, s_h, a_h, s_{h+1})$ do not proliferate much during a single stage, so $B$ becomes significantly smaller than $H \times S \times A \times S$. The auxiliary space complexity is $\approx O(S(HA+H+A)+B)$.

\section{Experimental setting}
We demonstrate the efficacy of our approaches in two non-stationary MDP settings, RandomMDP and BDCL.

\subsection{MDPs}
Inspired by the need for robust testing on different environment types for non-stationary RL, we introduce the pseudorandomly generated environment RandomMDP. In brief, each action deterministically transitions a state to a randomly picked next state and rewards for each state, action combination are randomly generated to be between 0 and 1. To introduce non-stationarity, a new \textit{target} MDP is randomly generated, and the available variation budget (decided by input parameters) is used to change the current MDP to the target. Once the target is reached, a new target is generated, and this process can repeat infinitely. Further detail of the generation process and input parameters can be found in Appendix \ref{sec:randommdp}.

In this study, we also focus on BDCL, an episodic MDP designed to be particularly challenging for exploration. The environment presents two types of non-stationary dynamics based on configuration: \textit{abrupt} and \textit{gradual} variations. We explain the details of BDCL in Appendix \ref{sec:bdcl}.

\subsection{Methodology}
We test our restart algorithms on RandomMDP with $A=5$, $S=5$, $H=5$, $T=50,000$ and varying input parameters to simulate different types of variation. We also test on gradual and abrupt BDCL $A=5$, $H=5$, $T=100,000$, and fail probability $0.02$. In the abrupt setting, changes happen at every $1,001$ episodes. We iterate each trial $5$ times.

We test the following combinations of algorithms: (1) \adaptiver + \partialr + \rqucb (\adaparrqucb) and (2) \selectiver + \randomized (\selectiverandomized). As baselines, we compare them with \rqucb, a random policy, and the optimal policy. Following the proof by \citet{mao2022modelfreenonstationaryrlnearoptimal}, $b_\Delta$ is removed (set to zero) from the update rule. In \selectiverandomized, we set $20$ ensembles, inflation coefficient $\kappa = 1$, and $n_0=\frac{1}{4}$ prior transitions. 

Since our main focus is the comparison against \rqucb, we performed hyperparameter tuning to find the best value for the probability hyperparameter $\delta$. $\delta \in [0,1]$ represents a probability required for the theoretical guarantees and impressive dynamic regret bound---the smaller $\delta$, the more optimism is added but the more likely the bounds are to hold. However, in practice, $\delta$ may be as large as $2$; then we set $\iota = \log\left(\frac{2}{\delta}\right)$ to zero on line 8 of Algorithm \ref{alg:rqucb}. Indeed, although this breaks the proofs in \citet{mao2022modelfreenonstationaryrlnearoptimal}, we find empirically that $\delta=2$ is the optimal setting of this hyperparameter.

\section{Results and discussion}

\begin{figure}[htbp]
\includegraphics[width=0.33\textwidth]{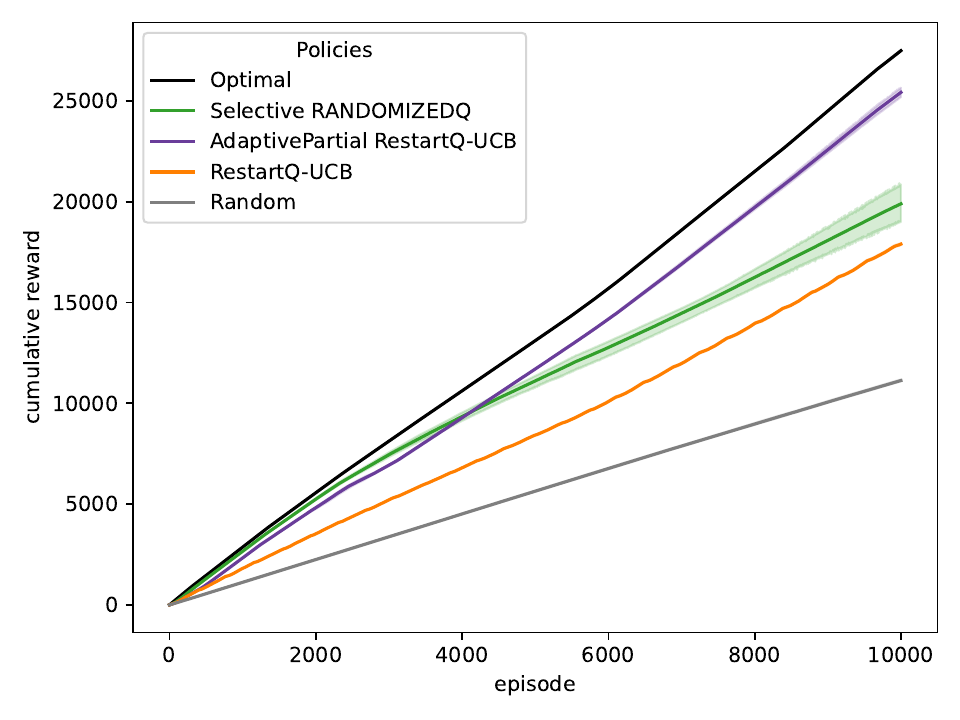}
\includegraphics[width=0.33\textwidth]{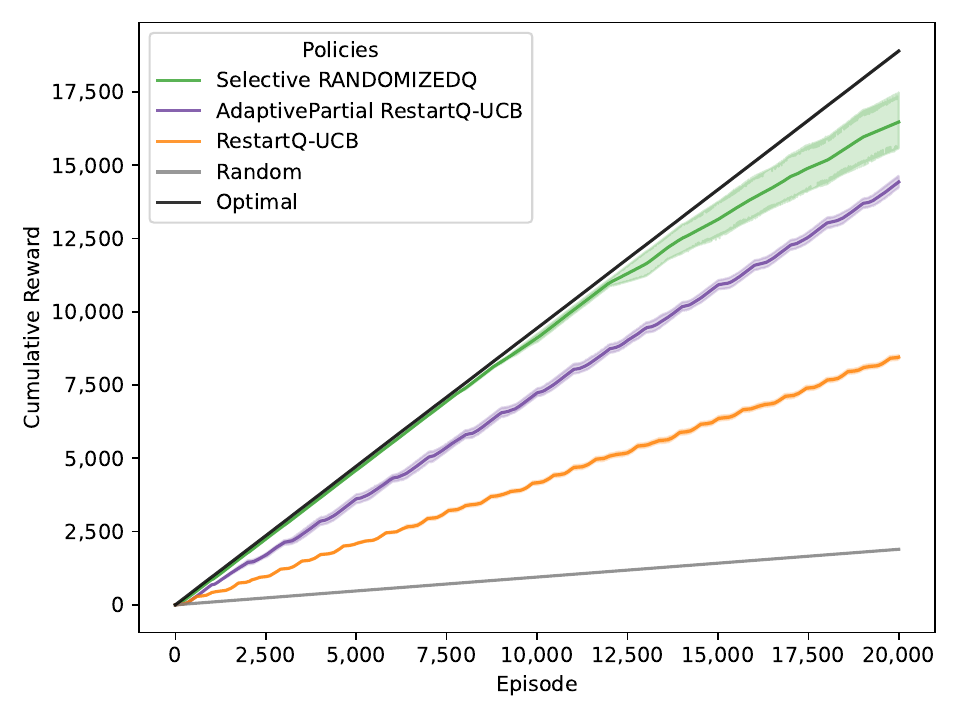}
\includegraphics[width=0.33\textwidth]{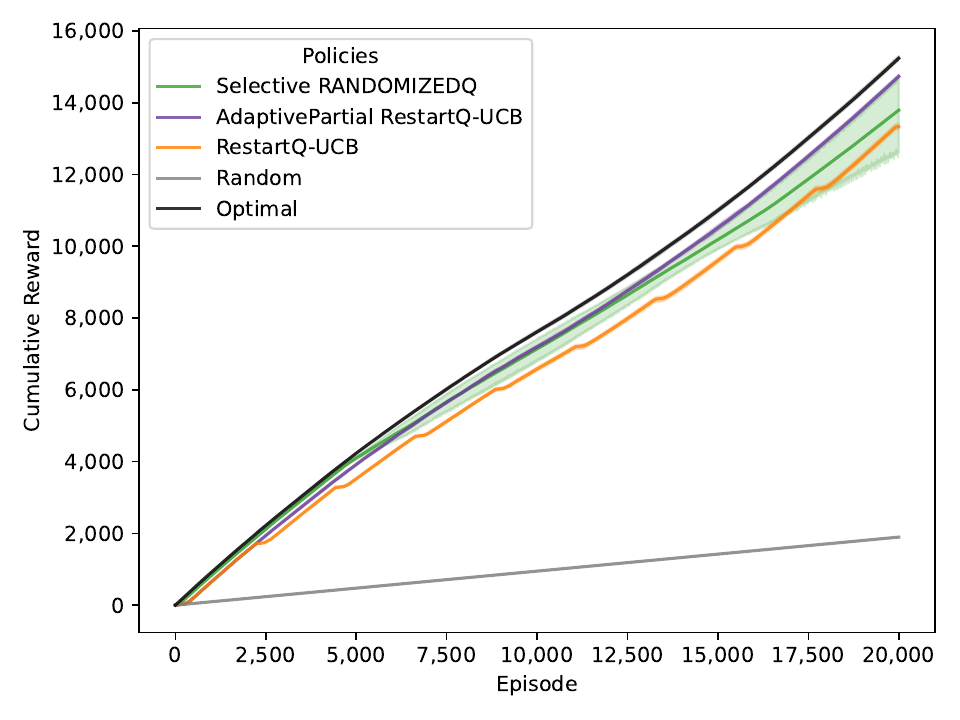}
\caption{From the left, each plot corresponds to RandomMDP, abrupt BDCL, gradual BDCL. In RandomMDP, \rqucb with adaptive and partial restarts achieves near-optimal total reward, and shows great improvement over base \rqucb. In BDCL environments, \rqucb with adaptive and partial restarts, as well as \selectiverandomized show an improved performance compared to base \rqucb. Notably, \selectiverandomized has near-zero dynamic regret in abrupt BDCL at episode 7,500, showing the promise of this approach.}

\label{fig:BDCL_total_reward}
\end{figure}

Figure \ref{fig:BDCL_total_reward} shows the total reward of the algorithms on RandomMDP, abrupt, and gradual BDCL. Overall, \adaparrqucb and \selectiverandomized achieve a significantly higher reward than \rqucb not only in RandomMDP but also in BDCL, in which an agent observes sparse reward signals and may struggle with exploration. These results demonstrate that our approaches successfully address the restart inefficiencies that we discuss in Section \ref{sec:existing_limitations}.

\paragraph{Partial and adaptive restarts} Partial and adaptive restarts combined in \rqucb prove to be a substantial improvement over base \rqucb. Partial restarts function properly in environments regardless as to whether the non-stationarity stems from changes in reward, transition probabilities, or both. Adaptive restarts effectively handle both abrupt and gradual change types. Figure \ref{fig:BDCL_total_reward} shows that when combined in a BDCL environment with abrupt changes, adaptive partial restarts decrease the dynamic regret of \rqucb by 45\%. In the RandomMDP, it decreases the dynamic regret by 74\%, showing great improvement in restarts. We observe that the large difference in reward is caused by \rqucb restarting very often, usually at times when \textit{not} restarting would still allow the agent to gain the maximum reward per episode. Adaptive restarts triggers very sparingly in comparison.

\paragraph{Selective restarts} Selective restarts demonstrate near-optimal performance in the three environments, especially early in the runtime, significantly reducing dynamic regret. The most notable example is shown in the middle of Figure \ref{fig:BDCL_total_reward}. Selective restarts maintain almost optimal rewards in the first ten thousands of episodes, and similar trends are observable in RandomMDP and gradual BDCL in the first few thousands of episodes. In abrupt BDCL, selective restarts achieved $91\%$ less dynamic regret relative to \rqucb. This is possible because of the algorithmic fit of selective restarts with agile stationary algorithms.

\subsection{Limitations}
The core limitation of this work is that our proposed methods are tested in only two settings and does not guarantee the performance of our approaches in every MDP. However, we believe that this work is an insightful step towards that for both theorists and practitioners. Below, we discuss in more depth how our approaches could be improved and how they may perform in environments that we did not examine.

\paragraph{Partial restarts} Knowledge of $\Delta_p$ and $\Delta_r$ is a strong assumption in practice, and in many real-world scenarios this value is unknown. When implementing partial restarts without knowledge of variation budget, it must either be estimated through repeated sampling (which will likely have poor theoretical performance), or the budgets can be entered as values much higher than expected, so that the true budget shouldn't exceed them. In practice, this will be much closer to full restarts than the tightly-bounded partial restarts shown here. Future work could investigate how both approaches manage in different settings.

\paragraph{Adaptive restarts} While adaptive restarts function well in the RandomMDP and BDCL environments shown here, worst-case environments can be constructed where these adaptive restarts will perform worse than scheduled ones. When implementing them, it should be tested if adaptive restarts will trigger at expected timings, and likely a combination of spaced-out scheduled restarts and adaptive restarts will give good performance while maintaining some degree of asymptotic guarantee.

\paragraph{Selective restarts} Although we propose selective restarts as a new restart framework, the derivation of the update amount found in Equation \ref{eqn:deltaQ} is still heuristic and needs a proof-based foundation. Moreover, we discover that selective restarts tend to start accumulating dynamic regret when the episode reaches a certain count, so future work should develop a new version derived from theoretical analysis.

\section{Conclusion}
In this paper, we focus on improving restart properties inspired by \rqucb. We propose partial, adaptive, and selective restarts, all of which successfully address the inherent problems of \rqucb (complete forgetting and inflexible restart timings) and shed light on achieving even quicker convergence in combination with other stationary algorithms. This work demonstrates that our approaches successfully elevate the efficacy of theoretically robust algorithms in experimental settings, shortening the gap between theory and practice that is prevalent in RL research.

\paragraph{Future work} These results give rise to some promising future directions. Adaptive restarts can almost certainly be improved by taking a more theoretical approach to their design. It currently functions using a heuristic that works well in these environments, but it would not detect a change in a setting where the change does not decrease the agent's rate of reward. One notable finding in selective restarts is that \selectiverandomized starts diverging from the optimal policy after some thousands of episodes and widens the confidence interval. Future improvements should focus on making its performance more robust by revisiting the update trigger condition and deriving a $\Delta Q$ that guarantees asymptotic performance (Equation \ref{eqn:deltaQ}). Nonetheless, the near-optimal performance early in the horizon implies positive possibilities for future improvements. Additionally, the asymptotic performance of these approaches could be found, and modifications may need to be made for more robust performance in general non-stationary MDPs. One can also apply our \textit{restart wrappers} to other stationary algorithms. Since our approaches offer only simple, outer modifications to the algorithms, we believe this is a hopeful direction. Another direction is to integrate a future well-tailored stationary algorithm with these new restart frameworks to achieve potentially lower dynamic regret.

\section*{Acknowledgments}
Aviva Prins was supported by Red Cell Partners. We would like to thank the students of REU-CAAR, along with Dr. William Gasarch, for their help and support during our research.

\bibliographystyle{IEEEtranN}
\bibliography{references}

\clearpage

\appendix

\section{Proof of Lemma 1} \label{sec: proof of lemma1}

\begin{manuallemma}{1}
    For any triple $(s,a,h)$ and any episodes $k_1, k_2 \in [K]$, it holds that $$ \left| Q^{k_2, \star}_h(s,a)  - Q^{k_1, \star}_h(s,a) \right| \leq \Delta_r + \frac{1}{2}\Delta_p\min_{k \in k_1, k_2}\left[ \max_{s,a,h' > h}\left[Q^{k, \star}_{h'}(s,a)\right]\right].$$
\end{manuallemma}

\begin{proof}
First, let us revise some notation. Let $Q^{k,\star}_h(s,a)$ be the $Q$-score of state $s$ and action $a$ given the optimal policy $\pi^\star$ at episode $k$ and timestep $h$. Define $V^{k,\star}_h(s, a)$ similarly. Define $\mathbbm{E}
\Bigl[
V^{k,\star}_{h+1}(s)
\Bigl] := \sum_{s' \in S} \Bigl[ 
P^k_h(s'|s,a)V^{k,\star}_{h+1}(s') \Bigl]$.
Let $\Delta_r$ and $\Delta_p$ be the total reward and probability variation budgets, respectively.
Let $\Delta_{r,h}$ and $\Delta_{p,h}$ be the variation budgets confined to a single timestep, such that
\[
\sum^H_{h=1}\Delta_{r,h} = \Delta_r\,,\;
\sum^H_{h=1}\Delta_{p,h} = \Delta_p.
\]
Finally, let $\alpha_h = \sum^H_{h'=h}
\Bigg[
\Delta_{r,h'}
+
\frac{1}{2}\Delta_{p,h'}\min_{k \in k_1, k_2}\left[ \max_{s,a}Q^{k,\star}_{h'+1}(s,a) \right]
\Bigg]$.

Our proof will be via backwards induction on $h$. This generalizes to all epochs, so $d$ will be omitted in notation. Without loss of generality, assume $k_2 > k_1$. If this is not the case, flipping the order of sums from $k_1$ to $k_2$ will still hold. Lastly, if $k_2 = k_1$, the proof is trivial. 

\textbf{Base case: $(h = H)$}

\begin{align*}
\left|Q^{k_2,\star}_H(s,a) - Q^{k_1,\star}_H(s,a) \right|&= 
\left|
r^{k_2}_H(s,a) + \mathbb{E}
\Bigg[
V^{k_2,\star}_{H+1}(s')
\Bigg]
-\:\, r^{k_1}_H(s,a) - \mathbb{E}
\Bigg[
V^{{k_1},\star}_{H+1}(s')
\Bigg]
\right|
\\
&= \left| r^{k_2}_H(s,a) - r^{k_1}_H(s,a) 
\right|
\\
&\leq
\sum^{k_2-1}_{k=k_1}\left| r^{k+1}_H(s,a) - r^{k}_H(s,a) 
\right|
\\
&\leq
\sum^{K-1}_{k=1}\left| r^{k+1}_H(s,a) - r^{k}_H(s,a) 
\right|
\\
&\leq \Delta_{r,H}
\end{align*}
 
\textbf{Inductive Hypothesis:}

\[
Q^{k_2,\star}_{h+1}(s,a) - Q^{k_1,\star}_{h+1}(s,a) \leq \alpha_{h+1}
\]

\textbf{Inductive Step:}

\begin{align*}
Q^{k_2,\star}_h(s,a) - Q^{k_1,\star}_h(s,a) 
&= 
r^{k_2}_h(s,a) 
-\:\, r^{k_1}_h(s,a) 
+ \mathbbm{E}
\Bigg[
V^{k_2,\star}_{h+1}(s')
\Bigg]
- \mathbbm{E}
\Bigg[
V^{k_1,\star}_{h+1}(s')
\Bigg]
\\
&\leq
\sum^{k_2-1}_{k=k_1} \left[ r^{k+1}_h(s,a) - r^{k}_h(s,a) 
\right]
 + 
\mathbbm{E}
\Bigg[
V^{k_2,\star}_{h+1}(s')
\Bigg]
- \mathbbm{E}
\Bigg[
V^{k_1,\star}_{h+1}(s')
\Bigg]
\\
&\leq
\sum^{K-1}_{k=1}
\left[ r^{k+1}_h(s,a) - r^{k}_h(s,a) 
\right]
 + 
\mathbbm{E}
\Bigg[
V^{k_2,\star}_{h+1}(s')
\Bigg]
- \mathbbm{E}
\Bigg[
V^{k_1,\star}_{h+1}(s')
\Bigg]
\\
&\leq
\Delta_{r,h} + 
\mathbbm{E}
\Bigg[
V^{k_2,\star}_{h+1}(s')
\Bigg]
- \mathbbm{E}
\Bigg[
V^{k_1,\star}_{h+1}(s')
\Bigg]
\\
&=
\Delta_{r,h} +
\sum_{s' \in S} 
P^{k_2}_h(s'|s,a)V^{k_2,\star}_{h+1}(s')
-
P^{k_1}_h(s'|s,a)V^{k_1,\star}_{h+1}(s')
\\
&= \Delta_{r,h} +
\sum_{s' \in S} 
P^{k_2}_h(s'|s,a)Q^{k_2,\star}_{h+1}(s',\pi^{k_2,\star}_{h+1}(s'))
-
P^{k_1}_h(s'|s,a)V^{k_1,\star}_{h+1}(s') 
\\
& =
\Delta_{r,h} + \sum_{s' \in S} 
P^{k_2}_h(s'|s,a)Q^{k_2,\star}_{h+1}(s',\pi^{k_2,\star}_{h+1}(s'))
-
P^{k_1}_h(s'|s,a)Q^{k_1,\star}_{h+1}(s',\pi^{k_1,\star}_{h+1}(s'))
\\
& \leq
\Delta_{r,h} + \sum_{s' \in S} 
P^{k_2}_h(s'|s,a)Q^{k_2,\star}_{h+1}(s',\pi^{k_2,\star}_{h+1}(s'))
-
P^{k_1}_h(s'|s,a)Q^{k_1,\star}_{h+1}(s',\pi^{k_2,\star}_{h+1}(s'))
\\
&\quad \textbf{By the I.H.}
\\
& \leq
\Delta_{r,h} + \sum_{s' \in S} 
P^{k_2}_h(s'|s,a)(Q^{k_1,\star}_{h+1}(s',\pi^{k_2,\star}_{h+1}(s')) + \alpha_{h+1})
-
P^{k_1}_h(s'|s,a)Q^{k_1,\star}_{h+1}(s',\pi^{k_2,\star}_{h+1}(s')) 
\\
&=
\Delta_{r,h} +
\sum_{s' \in S} 
(P^{k_2}_h(s'|s,a) - P^{k_1}_h(s'|s,a))Q^{k_1,\star}_{h+1}(s',\pi^{k_2,\star}_{h+1}(s')) + P^{k_2}_h(s'|s,a)\alpha_{h+1}
\\
&=
\Delta_{r,h} + \alpha_{h+1} +
\sum_{s' \in S}  
(P^{k_2}_h(s'|s,a) - P^{k_1}_h(s'|s,a))Q^{k_1,\star}_{h+1}(s',\pi^{k_2,\star}_{h+1}(s'))
\end{align*}

We can find this additional result by applying the I.H. to the right instead of the left side of the sum:
\begin{align*}
&\Delta_{r,h} + \sum_{s' \in S} 
P^{k_2}_h(s'|s,a)Q^{k_2,\star}_{h+1}(s',\pi^{k_2,\star}_{h+1}(s'))
-
P^{k_1}_h(s'|s,a)Q^{k_1,\star}_{h+1}(s',\pi^{k_2,\star}_{h+1}(s'))
\\
&\quad \textbf{By the I.H.}
\\
& \leq
\Delta_{r,h} + \sum_{s' \in S} 
P^{k_2}_h(s'|s,a)Q^{k_2,\star}_{h+1}(s',\pi^{k_2,\star}_{h+1}(s'))
-
P^{k_1}_h(s'|s,a)(Q^{k_2,\star}_{h+1}(s',\pi^{k_2,\star}_{h+1}(s')) - \alpha_{h+1})
\\
&=
\Delta_{r,h} +
\sum_{s' \in S} 
(P^{k_2}_h(s'|s,a) - P^{k_1}_h(s'|s,a))Q^{k_2,\star}_{h+1}(s',\pi^{k_2,\star}_{h+1}(s')) + P^{k_1}_h(s'|s,a)\alpha_{h+1}
\\
&=
\Delta_{r,h} + \alpha_{h+1} +
\sum_{s' \in S}  
(P^{k_2}_h(s'|s,a) - P^{k_1}_h(s'|s,a))Q^{k_2,\star}_{h+1}(s',\pi^{k_2,\star}_{h+1}(s'))
\end{align*}

We now bound $\sum_{s' \in S} \Bigl[ 
(P^{k_2}_h(s'|s,a) - P^{k_1}_h(s'|s,a))Q^{k_1,\star}_{h+1}(s',\pi^{k_2,\star}_{h+1}(s'))
\Bigl]$. (We want to show $\leq \frac{1}{2}\Delta_{p,h}\max_{s,a}Q^{k_1,\star}_{h+1}(s,a)$).

\textbf{Note}: The following steps are also used to bound: 

$\sum_{s' \in S} \Bigl[ 
(P^{k_2}_h(s'|s,a) - P^{k_1}_h(s'|s,a))Q^{k_2,\star}_{h+1}(s',\pi^{k_2,\star}_{h+1}(s'))
\Bigl]$.

Because $P^{k_2}_h(\cdot|s,a)$ and $P^{k_1}_h(\cdot|s,a)$ are probability vectors, the following is true:
\[
\sum_{s' \in S}
P^{k_2}_h(s'|s,a) = 1 = 
\sum_{s' \in S} P^{k_1}_h(s'|s,a)
\]

\[
\rightarrow \sum_{s' \in S} \Bigg[ 
P^{k_2}_h(s'|s,a) - P^{k_1}_h(s'|s,a) \Bigg]
=
0
\]

By the definition of $\Delta_{p,h}$, the following is also true:

\[
\sum_{s' \in S} \Bigg[ 
\Bigl| P^{k_2}_h(s'|s,a) - P^{k_1}_h(s'|s,a) 
\Bigl|
\Bigg]
\leq \Delta_{p,h}
\]

\vspace{1em}

For ease of notation going forward, let $\theta_i = P^{k_2}_h(s_i|s,a) - P^{k_1}_h(s_i|s,a)$, $Q_i = Q^{k_1,\star}_{h+1}(s_i,\pi^{{k_2},\star}_{h+1})$, and let $n = |S|$, the number of states. Therefore, the above equations and inequality become:

\begin{align*}
\sum^{n}_{i =1}
\Bigl| \theta_i 
\Bigl|
&\leq \Delta_{p,h}
\\
\sum^{n}_{i =1}
\theta_i 
&= 0
\end{align*}

\[
\sum_{s' \in S} \Bigl[ 
(P^{k_2}_h(s'|s,a) - P^{k_1}_h(s'|s,a))Q^{{k_1},\star}_{h+1}(s',\pi^{{k_2},\star}_{h+1})
\Bigl]
= \sum^{n}_{i =1}
\theta_iQ_i
\]

We will show the sum of all positive $\theta_i$ is less than or equal to $\frac{1}{2}\Delta_{p,h}$. Since the sum of the elements in $\theta_i$ is equal to $0$,

\[
\sum_{i:\theta_i > 0} \theta_i = - \sum_{i:\theta_i < 0} \theta_i
\]

Therefore,

\[
\sum_{i=1}^n |\theta_i| = \sum_{i:\theta_i > 0} \theta_i - \sum_{i:\theta_i < 0} \theta_i = 2 \sum_{i:\theta_i > 0} \theta_i
\]

Since the sum $\sum_{i=1}^n |\theta_i|$ is bounded by $\Delta_{p, h}$,

\begin{align*}
2 \sum_{i:\theta_i > 0} \theta_i &\leq \Delta_{p,h}
\\
\sum_{i:\theta_i > 0} \theta_i &\leq \frac{1}{2} \Delta_{p,h}
\end{align*}

Now we return to our sum $\sum^{n}_{i =1}
\theta_iQ_i$.

Since all $Q_i \geq 0$:

\begin{align*}
\sum_{s' \in S} \Bigl[ 
(P^{k_2}_h(s'|s,a) - P^{k_1}_h(s'|s,a))Q^{{k_1},\star}_{h+1}(s',\pi^{{k_2},\star}_{h+1})
\Bigl]
& = \sum^{n}_{i =1}
\theta_iQ_i
\\
&\leq
\sum^{n}_{i=1:\theta_i > 0}
\theta_iQ_i 
\\
&\leq 
\sum^{n}_{i=1:\theta_i > 0}
\theta_i\max_s(Q_s)
\\
&\leq \frac{1}{2} \Delta_{p,h} \max_s(Q_s)
\\
&= \frac{1}{2} \Delta_{p,h}\max_s(Q^{k_1,\star}_{h+1}(s,\pi^{{k_2},\star}_{h+1}(s)))
\\
&\leq \frac{1}{2} \Delta_{p,h}\max_{s,a}(Q^{k_1,\star}_{h+1}(s,a))
\end{align*}

\vspace{1em}

Similarly,

\[
\sum_{s' \in S} \Bigl[ 
(P^{k_2}_h(s'|s,a) - P^{k_1}_h(s'|s,a))Q^{{k_2},\star}_{h+1}(s',\pi^{{k_2},\star}_{h+1})
\Bigl]
\leq
\frac{1}{2} \Delta_{p,h}\max_{s,a}(Q^{k_2,\star}_{h+1}(s,a))
\]
Combining all above steps,

\begin{align*}
Q^{k_2,\star}_h(s,a) - Q^{k_1,\star}_h(s,a)  &\leq \Delta_{r,h} + \alpha_{h+1} +
\sum_{s' \in S} \Bigg[ 
(P^{k_2}_h(s'|s,a) - P^{k_1}_h(s'|s,a))Q^{k_1,\star}_{h+1}(s',\pi^{{k_2},\star}_{h+1})
\Bigg]
\\
&\leq \Delta_{r,h} + \alpha_{h+1} + \frac{1}{2} \Delta_{p,h}\max_{s,a}(Q^{k_1,\star}_{h+1}(s,a))
\\
&=
\Delta_{r,h} +
\sum^H_{h'=h+1}
\Bigg[
\Delta_{r,h'}
+
\frac{1}{2}\Delta_{p,h'}\max_{s,a}Q^{k_1,\star}_{h'+1}(s,a)
\Bigg] + \frac{1}{2} \Delta_{p,h}\max_{s,a}(Q^{k_1,\star}_{h+1}(s,a))
\\
&= \sum^H_{h'=h}
\Bigg[
\Delta_{r,h'}
+
\frac{1}{2}\Delta_{p,h'}\max_{s,a}Q^{k_1,\star}_{h'+1}(s,a)
\Bigg]
\end{align*}

Through a similar process, it can be shown that:

\[
Q^{k_2,\star}_h(s,a) - Q^{k_1,\star}_h(s,a)  \leq \sum^H_{h'=h}
\Bigg[
\Delta_{r,h'}
+
\frac{1}{2}\Delta_{p,h'}\max_{s,a}Q^{k_2,\star}_{h'+1}(s,a)
\Bigg]
\]

Combining the two results:

\begin{align*}
Q^{k_2,\star}_h(s,a) - Q^{k_1,\star}_h(s,a)  
&\leq
\sum^H_{h'=h}
\Bigg[
\Delta_{r,h'}
+
\frac{1}{2}\Delta_{p,h'}\min_{k \in k_1, k_2}\left[ \max_{s,a}Q^{k,\star}_{h'+1}(s,a) \right]
\Bigg]
\\
& = \alpha_h
\\
&\leq
\Delta_r
+
\frac{1}{2}\Delta_p\min_{k \in k_1, k_2}\left[ \max_{s,a, h'>h}Q^{k,\star}_{h'}(s,a) \right]
\end{align*}

Finally, because of our assumption (without loss of generality) that $k_2 > k_1$, bounding $Q^{k_1,\star}_h(s,a) - Q^{k_2,\star}_h(s,a)$ will yield the same result. Therefore:

\[
\left| Q^{k_2,\star}_h(s,a) - Q^{k_1,\star}_h(s,a) 
\right|
\leq
\Delta_r
+
\frac{1}{2}\Delta_p\min_{k \in k_1, k_2}\left[ \max_{s,a, h'>h}Q^{k,\star}_{h'}(s,a) \right]
\]

\end{proof}

\section{\adaptiver Algorithm}
\label{sec:adaptivealgorithm}

The following algorithm is \rqucb modified to use adaptive restarts, described in \ref{sec: adaptive restarts}. For clarity, all variables using $\mathtt{snake\_case}$ are newly introduced for adaptive restarts.

\begin{algorithm}[H] \label{alg:adarqucb}
\caption{\rqucb (Hoeffding) with Adaptive Restarts}

\While{$\mathtt{restart} == \texttt{True}$}
    {\textbf{Initialize:} $V_h(s) \gets H - h + 1$, $Q_h(s,a) \gets H - h + 1$,
    $N_h(s,a) \gets 0$, $\check{N}_h(s,a) \gets 0$, 
    $\check{r}_h(s,a) \gets 0$, $\check{v}_h(s,a) \gets 0$, 
    $\check{\mu}_h(s,a) \gets 0$, $\check{\sigma}_h(s,a) \gets 0$,
    $\mu^{\text{ref}}_h(s,a) \gets 0$, $\sigma^{\text{ref}}_h(s,a) \gets 0$, 
    $V^{\text{ref}}_h(s) \gets H$, for all $(s,a,h) \in \mathcal{S} \times \mathcal{A} \times [H]$.
    
    \textbf{Initialize (restart vars):} 
    $\mathtt{W} \gets 0$, $r_L \gets 0$, $r_B \gets 0$, $r_C \gets 0$, 
    $\mathtt{trueCount} \gets 0$, $\mathtt{nonCount} \gets 0$, $\mathtt{lastReset} \gets 0$.
    $\mathtt{rewardHistory} \gets [\,]$.
    
    \For{episode $k \gets k + 1$ to $T$}
        {\textbf{observe} $s_1$, \; $\mathtt{episodeReward} \gets 0$;
        
        \For{step $h \gets 1$ to $H$}
        {
            \textbf{Take} action $a_h \gets \arg\max_a Q_h(s_h,a)$;
            \textbf{Receive} reward $R_h(s_h,a_h)$ and observe $s_{h+1}$;
            
            $\check{r}_h(s_h,a_h) \gets \check{r}_h(s_h,a_h) + R_h(s_h,a_h)$,
            $\check{v}_h(s_h,a_h) \gets \check{v}_h(s_h,a_h) + V_{h+1}(s_{h+1})$,
            $\mathtt{episodeReward} \gets \mathtt{episodeReward} + R_h(s_h,a_h)$;
            
            $N_h(s_h,a_h) \gets N_h(s_h,a_h) + 1$, 
            $\check{N}_h(s_h,a_h) \gets \check{N}_h(s_h,a_h) + 1$;
            
            \If{$N_h(s_h,a_h) \in \mathcal{L}$}{
                $b_h \gets \sqrt{\tfrac{H^2}{\check{N}_h(s_h,a_h)} \, \iota}
                         + \sqrt{\tfrac{1}{\check{N}_h(s_h,a_h)} \, \iota}$,
                $b_\Delta \gets \Delta_r^{(d)} + H \cdot \Delta_p^{(d)}$;
                
                $Q_h(s_h,a_h) \gets \min \Bigl\{
                       \tfrac{\check{r}_h(s_h,a_h)}{\check{N}_h(s_h,a_h)}
                       + \tfrac{\check{v}_h(s_h,a_h)}{\check{N}_h(s_h,a_h)}
                       + b_h + 2b_\Delta, 
                       Q_h(s_h,a_h) \Bigr\}$;
                
                $V_h(s_h) \gets \max_a Q_h(s_h,a)$;
                
                \If{optimal action at $(s_h,h)$ unchanged}{
                    $\mathtt{nonCount} \gets \mathtt{nonCount} + 1$;
                }\Else{
                    $\mathtt{trueCount} \gets \mathtt{trueCount} + 1$;
                }
                
                \If{$\mathtt{trueCount} \geq H^2$}{
                    $\mathtt{trueCount}, \mathtt{nonCount} \gets 0,0$;
                    $\mathtt{lastReset} \gets k$; \Comment{still learning}
                }
                \If{$\mathtt{nonCount} \geq H^2$}{
                    $W \gets k - \mathtt{lastReset}$; \Comment{learning done}
                }
                
                $\check{N}_h(s_h,a_h) \gets 0$, 
                $\check{r}_h(s_h,a_h) \gets 0$, 
                $\check{v}_h(s_h,a_h) \gets 0$;
            }
        }
        
        Append $\mathtt{episodeReward}$ to $\mathtt{rewardHistory}$;
        
        \If{$\mathtt{W} > 0$}{
            $r_L \gets \sum_{j=1}^W \mathtt{rewardHistory}[j]$; \Comment{learning reward}
            
            $r_C \gets \sum_{j=k-W+1}^k \mathtt{rewardHistory}[j]$; \Comment{current reward}
            
            $r_B \gets \max(r_B, r_C)$; \Comment{best reward so far}
            
            $R_{\text{no}} \gets r_C \cdot \tfrac{T-t}{HW}$, \;
            $R_{\text{yes}} \gets r_L + r_B \cdot \Bigl(\tfrac{T-t}{HW} - 1\Bigr)$;
            
            \If{$R_{\text{no}} < R_{\text{yes}}$}{
                \textbf{Restart: reinitialize all variables as at start of epoch.}
            }
        }
        }
    }
\end{algorithm}

\section{\selectiver Algorithm} \label{sec:selectivealgorithm}
The following is the algorithm of our \selectiver. In the following pseudocode, only trajectory $\mathcal{T}$ and the trajectory dictionary $T$ are initialized for \selectiver. We used a dictionary to store the trajectory, which significantly reduces the space complexity over using an $H \times S \times A \times S$ matrix, where the dimensions correspond to $(h, s_h, a_{h}, s_{h+1})$. The time complexity of the function $f_\mathcal{T}$ that retrieves a set of $(h', s_{h'}, a_{h'}, s_{h'+1})$ quadruples from $T$ may vary, but in our algorithm, it is $O(B)$. 

\RestyleAlgo{ruled}
\begin{algorithm}[H]
\caption{SelectiveRestarts}\label{alg:selective}
\Input {Learning stages: $\mathcal{L}$; Visitation count: $N_h(s, a)$; Set of arbitrary visitation counts: $\hat{N}_h(s, a) = \set{\hat{N}_{1, h}(s, a), \hat{N}_{2, h}(s, a), \dots, \hat{N}_{n, h}(s, a)}$}
\Require {Policy update function: $f_\pi$; Value update function: $f_V$; Trajectory calculation function: $f_\mathcal{T}$; Current undiscounted step-wise Bellman update: $U^{k}_h(s, a)$; Bellman update at past episode $k_0$: $U^{k_0}_h(s, a)$; Upper bound of $\left| Q^{k_1, \star}_h(s, a) - Q^{k_2, \star}_h(s, a) \right|$ for $0 \le k_1 < k_2 \le K$: $\beta_h(s, a)$; Standard deviation of $Q$ along actions: $\sigma(Q_h(s, \cdot))$;}
\Initialize {Trajectory $\mathcal{T}_h(s, a) \leftarrow \phi$; Trajectory dictionary: $T$}
\For{$episode ~ m\gets1$ \KwTo $M$}{
    Observe the initial state $s^k_1$;
    
  \For{step $h\gets1$ \KwTo $H$}{
    Take action $a^m_h = \pi^{m}_h(s^m_h)$;
    Observe $s^m_{h+1}$; Receive reward $r_h^m(s^m_h, a^m_h)$

    $T(h, s^m_h, a^m_h, s^m_{h+1}) \gets 1$
    
    \tcp{Execute the base policy update}
    $f_\pi(s^m_h, a^m_h, s^m_{h+1}, r^m_h(s^m_h, a^m_h))$;
    
    \tcp{Calculate undiscounted step-wise Bellman update}
    $U^{m}_h(s^{m}_h, a^{m}_h) = r^{m}_h(s^{m}_h, a^{m}_h) + V^{m}_{h+1}(s^{m}_{h+1})$
    
    \If{$\left| U^{m_0}_h(s^m_h, a^m_h) - U^{m}_h(s^{m}_h, a^{m}_h) \right| \ge \beta_h(s^m_h, a^m_h)$}{
    
        \tcp{Find the trajectory to $(h, s^m_h, a^m_h)$}
        $\mathcal{T}_h(s^m_h, a^m_h) \gets f_\mathcal{T} (T, h, s^m_h, a^m_h)$
        
        \For{$(h', s_{h'}, a_{h'}, s_{h'+1}) \in \mathcal{T}_h(s^m_h, a^m_h)$}
        {
            \tcp{Calculate update amount}
            $\Delta Q_{h'}(s_{h'}, a_{h'}) = \sign(U^m_h(s, a) - U^{m_0}_h(s, a)) \gamma \frac{1}{H - {h'}}\beta_{h'}(s_{h'}, a_{h'})$\\
            $Q_{h'}(s_{h'}, a_{h'}) \gets Q_{h'}(s_{h'}, a_{h'}) + \Delta Q_{h'}(s_{h'}, a_{h'})$\\
            $V_{h'}(s_{h'}) \gets f_V(Q)$
        }
        \tcp{If any, reset other visitation counts}
        $\hat{N}_{h}(s^m_h, a^m_h) \gets \set{0}^{n}$\\
        $U^{m_0}_h(s^m_h, a^m_h) \gets U^{m}_h(s^m_h, a^m_h)$
    }
    
    \If{$N_h(s^k_h, a^k_h) \in \mathcal{L}$}{
        \tcp{Reset the trajectory table}
        \For{$(h', s_{h'}, a_{h'}, s_{h'+1}) \in \mathcal{T}_h(s^m_h, a^m_h)$}{
            $T(h', s_{h'}, a_{h'}, s_{h'+1}) \gets 0$
        }
    }
  }
}
\end{algorithm}

\section{RandomMDP}
\label{sec:randommdp}

RandomMDP has many input parameters that influence generation. First, to define the state-action space and random seed, it requires \texttt{N\_STATES}, \texttt{N\_ACTIONS}, \texttt{EPISODE\_LENGTH}, and \texttt{MDP\_SEED}. To determine the total variation budgets, $\Delta_r$ and $\Delta_p$, it takes in \texttt{TOTAL\_DELTA\_R} and \texttt{TOTAL\_DELTA\_P} respectively. To determine how abrupt each kind of variation is, there is \texttt{DELTA\_R\_ABRUPTNESS} and \texttt{DELTA\_P\_ABRUPTNESS}, where each specifies the percent of episodes with 0 $\Delta_r$ and $\Delta_p$. \texttt{DELTA\_R\_BUDGET\_DISTRIBUTION} and \texttt{DELTA\_P\_BUDGET\_DISTRIBUTION} describe how the total budget is divided between episodes that have non-zero variation. "uniform" has every episode have the same amount of variation, while "linear" linearly increases the variation per episode starting from 0. \texttt{FAIL\_PROBABILITY} is the chance of transitioning to a random, non-target state. Lastly, \texttt{REWARD\_SPARSITY} is the percent of "low-reward" state-actions, where reward is sampled from [0, 0.2] instead of [0,1]. Below is the generation procedure for RandomMDP given these input parameters. 

In the simulations shown in this paper, RandomMDP was run with the following parameters: 
\texttt{EPISODE\_LENGTH: 5,
N\_STATES: 5,
N\_ACTIONS: 5,
TOTAL\_DELTA\_R: 5,
TOTAL\_DELTA\_P: 10,
DELTA\_R\_ABRUPTNESS: 0.999,
DELTA\_P\_ABRUPTNESS: 0.5,
DELTA\_P\_BUDGET\_DISTRIBUTION: uniform,
DELTA\_R\_BUDGET\_DISTRIBUTION: uniform,
FAIL\_PROBABILITY: 0.05,
REWARD\_SPARSITY: 0.8}

\vspace{1em}

\begin{algorithm}[H]
\caption{RandomMDP generation}\label{alg:randommdp}

\tcp{First generate the per-episode variation budget}

$\mathtt{num\_of\_eps\_with\_\Delta_r} \gets K(1-\mathtt{DELTA\_R\_ABRUPTNESS})$

$\mathtt{avg\_\Delta_r\_per\_ep} \gets \frac{\mathtt{TOTAL\_DELTA\_R}}{\mathtt{num\_of\_eps\_with\_\Delta_r}}$

Define $\Delta^i_r$ as the reward variation budget for episode $i$.

\For{$episode ~ i\gets1$ \KwTo $\mathtt{num\_of\_eps\_with\_\Delta_r}$}{
    Assign $\Delta^i_r$ based on the budget distribution defined in the config. If uniform, each gets the same. If linear, variation budget for each increases from 0 to $2(\mathtt{avg\_\Delta_r\_per\_ep})$.
}
\For{$episode ~ i \gets \mathtt{num\_of\_eps\_with\_\Delta_r}$ \KwTo $K$}{
    $\Delta^i_r \gets 0$
}

Based on $\mathtt{MDP\_SEED}$, randomly swap each $\Delta^i_r$ with another $\Delta^i_r$, so that the budget is distributed across all episodes.

Lines 1-9 using $\Delta_p$ in place of $\Delta_r$.

Generate $\mathtt{N\_STATES}$ states and $\mathtt{N\_ACTIONS}$ actions.

\For{$state ~ s \gets s_0$ \KwTo $s_S$}{
    \For{$action ~ a \gets a_0$ \KwTo $a_A$}{
        \For{$timestep ~ h \gets 1$ \KwTo $H$}{
            $P^1_h(s'|s,a) \gets 1-\mathtt{FAIL\_PROBABILITY}$, where $s'$ is randomly sampled from $S$.

            $P^1_h((\forall s \neq s'|s,a) \gets \frac{\mathtt{FAIL\_PROBABILITY}}{\mathtt{N\_STATES} - 1}$

            $r^1_h(s,a) \gets$ random value $\in [0,1]$, for $1-\mathtt{REWARD\_SPARSITY}$ \% of $(s,a,h)$ triples.

            $r^1_h(s,a) \gets$ random value $\in [0,0.2]$, for $\mathtt{REWARD\_SPARSITY}$ \% of $(s,a,h)$ triples.
        }
    }
}

$P^1$ and $r^1$ are the $\mathtt{current\_P}$ and $\mathtt{current\_r}$ of the MDP.

Repeat lines 9-17 to generate $\mathtt{target\_P}$ and $\mathtt{target\_r}$

$needed\Delta_r, \; needed\Delta_p \gets$ total $\Delta_r$ and $\Delta_p$ to go from the current to the target $r$ and $P$ functions, respectively.

$\alpha_r, \;\alpha_p \gets 0,\; 0$

\For{$episode ~ k \gets 2$ \KwTo $K$}{
    $\alpha_r \;+= \frac{\mathtt{needed\Delta_r}}{\Delta^{k-1}_r}$

    $\alpha_p \;+= \frac{\mathtt{needed\Delta_p}}{\Delta^{k-1}_p}$

    \If{$\alpha_r > 1$}{
        $\mathtt{current\_r} \gets \mathtt{target\_r}$
    
        Generate new $\mathtt{target\_r}$ using the same process as lines 16 and 17.

        $\alpha_r \gets 0$
    }

    \If{$\alpha_p > 1$}{
        $\mathtt{current\_P} \gets \mathtt{target\_P}$
        
        Generate new $\mathtt{target\_P}$ using the same process as lines 14 and 15.

        $\alpha_p \gets 0$
    }
    $\forall(h,s,a,s') P^k_h(s'|s,a) \gets \mathtt{current\_P}_h(s'|s,a)(1-\alpha_p) + \mathtt{target\_P}_h(s'|s,a)(\alpha_p)$

    $\forall(h,s,a) r^k_h(s,a) \gets \mathtt{current\_r}_h(s,a)(1-\alpha_p) + \mathtt{target\_r}_h(s,a)(\alpha_p)$
}
\end{algorithm}

\section{Bidirectional Diabolical Combination Locks (BDCL)}\label{sec:bdcl}
 BDCL starts with a fixed initial state and transitions to either \textit{lock1} or \textit{lock2} depending on an action. Once the agent reaches one of the locks, it needs to keep choosing correct actions $H$ times to stay in the lock and receive a final reward: $r_H(s_{\text{lock}}, a_{\text{correct}}) \in \set{0.25, 1.0}$, depending on which lock the agent is in. The locks return reward $0$ until the final reward. If the agent chooses an incorrect action, it goes to the \textit{sink} state, in which the agent can only receive a slight reward and cannot go back to a lock, no matter what action it takes. There are \textit{abrupt} and \textit{gradual} settings. In the abrupt setting, the final rewards of lock1 and lock2 switch at every some number of episodes. \textit{Fail probability} directs the agent to the sink from a lock with probability $p$ when it takes a correct action. In gradual BDCL, the transition probability from the initial state to a lock given a particular action changes at a linear scale at every episode, throughout the horizon.

\end{document}